\documentclass[letterpaper]{article} 
\usepackage{aaai24}  
\usepackage{times}  
\usepackage{helvet}  
\usepackage{courier}  
\usepackage[hyphens]{url}  
\usepackage{graphicx} 
\usepackage{natbib}  
\usepackage{caption} 
\usepackage{algorithm}
\usepackage{algorithmic}

\usepackage{newfloat}
\usepackage{listings}
\usepackage{amssymb}
\usepackage{amsmath}
\usepackage{paralist}
\usepackage[capitalise]{cleveref}
\usepackage[most]{tcolorbox}
\usepackage[legacycolonsymbols]{mathtools}
\newtheorem{definition}{Definition}

\usepackage{booktabs}
\graphicspath{ {./fig/} }

\newtcolorbox[auto counter,number within=subsection]{myBox}[3][]{
	arc=5mm,
	lower separated=false,
	fonttitle=\bfseries,
	colbacktitle=gray!10,
	coltitle=gray!50!black,
	enhanced,
	attach boxed title to top left={xshift=0.5cm,
		yshift=-2mm},
	colframe=gray!50!black,
	colback=gray!10,
	overlay={
		\node[draw=gray!50!black,thick,
		fill= gray!10,rounded corners=1mm, 
		yshift=0pt, 
		xshift=-0.5cm, 
		left, 
		text=gray!50!black,
		anchor=east,
		font=\bfseries] 
		at (frame.north east) {#3};},
	overlay={
		\node[draw=gray!50!black,thick,
		fill= gray!10,rounded corners=1mm, 
		yshift=0pt, 
		xshift=-0.5cm, 
		left, 
		text=gray!50!black,
		anchor=east,
		font=\bfseries] 
		at (frame.north east) {#3};},
	title=#2,#1}

\DeclareCaptionStyle{ruled}{labelfont=normalfont,labelsep=colon,strut=off} 
\floatstyle{ruled}
\newfloat{listing}{tb}{lst}{}
\floatname{listing}{Listing}
\title{Deployable Reinforcement Learning with Variable Control Rate}
\author {
    Dong Wang\textsuperscript{\rm 1},
    Giovanni Beltrame\textsuperscript{\rm 2}
}
\affiliations {
    \textsuperscript{\rm 1} dong-1.wang@polymtl.ca\\
    \textsuperscript{\rm 2} giovanni.beltrame@polymtl.ca\\
}

\begin{document}

\maketitle
\begin{abstract}
  Deploying controllers trained with Reinforcement Learning (RL) on real robots
  can be challenging: RL relies on agents' policies being modeled as Markov
  Decision Processes (MDPs), which assume an inherently discrete passage of
  time. The use of MDPs results in that nearly all RL-based control systems
  employ a fixed-rate control strategy with a period (or time step) typically
  chosen based on the developer's experience or specific characteristics of the
  application environment. Unfortunately, the system should be controlled at the
  highest, worst-case frequency to ensure stability, which can demand
  significant computational and energy resources and hinder the deployability of
  the controller on onboard hardware. Adhering to the principles of reactive
  programming, we surmise that applying control actions \emph{only when
    necessary} enables the use of simpler hardware and helps reduce energy
  consumption. We challenge the fixed frequency assumption by proposing a
  variant of RL with \emph{variable control rate}. In this approach, the policy
  decides the action the agent should take as well as the duration of the time
  step associated with that action. In our new setting, we expand Soft
  Actor-Critic (SAC) to compute the optimal policy with a variable control rate,
  introducing the Soft Elastic Actor-Critic (SEAC) algorithm. We show the
  efficacy of SEAC through a proof-of-concept simulation 
  driving an agent with Newtonian kinematics. Our experiments show higher
  average returns, shorter task completion times, and reduced computational
  resources when compared to fixed rate policies.
\end{abstract}

\section{Introduction}

\begin{figure}[h]
	\begin{center}
		\includegraphics[width=1.0\linewidth]{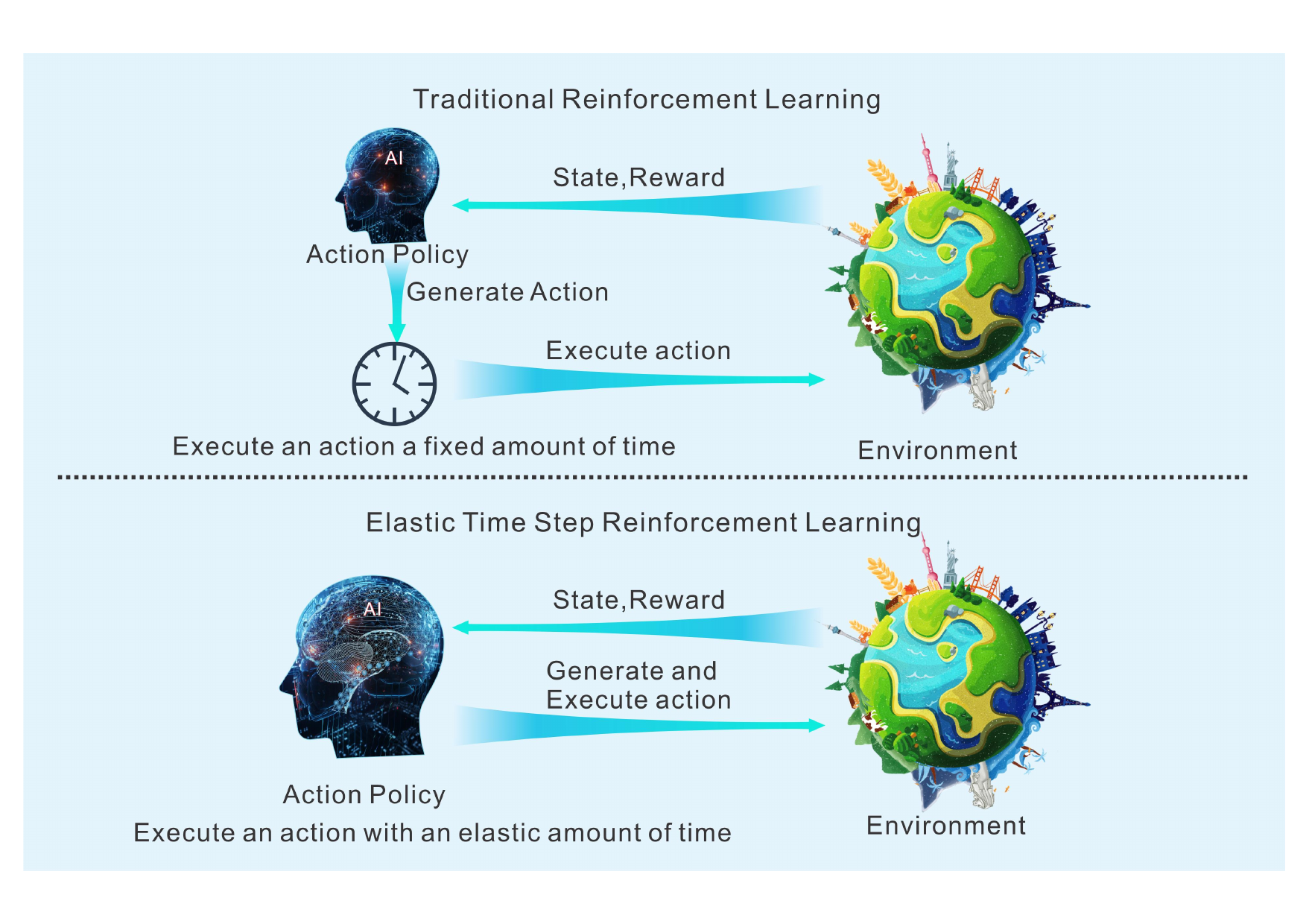}
	\end{center}
	\caption{Comparing Elastic Time Step Reinforcement Learning and Traditional Reinforcement Learning}
	\label{fig:overview}
\end{figure}

Temporal aspects of reinforcement learning (RL), such as the duration of the
execution of each action or the time needed for observations, are frequently
overlooked. This oversight arises from the foundational hypothesis of the
Markov Decision Process (MDP), which assumes the independence of each action
undertaken by the agent \citep{norris1998markov}. As depicted in the top
section of Figure~\ref{fig:overview}, conventional RL primarily focuses on
training an action policy, generally neglecting the intricacies of policy
implementation. Some prior research approached the problem by splitting 
their control algorithm into two distinct components 
\citep{williams2017information}: a learning part responsible for proposing an
action policy, and a control part responsible for implementing the policy
\citep{yang2018hierarchical,zanon2020safe,mahmood2018setting}. \footnote{Our 
	paper is presented at Deployable AI Workshop at AAAI-2024}

Translating action policies composed of discrete time steps into real-world
applications generally means using a fixed control rate (e.g., 10~Hz).
Practitioners typically choose the control rate based on their experience and
the specific needs of each application, often without considering adaptability
or responsiveness to changing environmental conditions. 
Imagine driving a car in a straight line in an extensive environment without
obstacles: very few control actions are needed. On the contrary, driving in
tight spaces with low tolerances and following complex paths can require a
higher control frequency. Setting the control frequency is set to a fixed value
requires it to be the worst-case (or an average if safety is not a concern)
scenario that allows controllability of the system.
In practical applications of reinforcement learning, especially in scenarios with
constrained onboard computer resources, maintaining a consistently high fixed
control rate can limit the availability of computing resources for other tasks
and significantly increase energy consumption.

Furthermore, the inherent inertia of physical systems cannot be ignored,
impacting the range of feasible actions. In such cases, an agent's control
actions are closely tied to factors like velocity and mass, leading to
considerably different outcomes when agents execute the same actions at
different control rates.

Hence, applying RL directly to real-world scenarios can be challenging when
the temporal dimension is not considered. The typical approach is to employ
\emph{a fast enough but fixed control rate that accommodates the worst-case
	scenario} for an application \citep{mahmood2018setting}, resulting in
suboptimal performance in most instances.

In this paper, we challenge RL's conventional fixed time step assumption, aiming
to formulate faster and more energy-efficient policies. Our approach seamlessly
integrates the temporal aspect into the learning process. This modification can
benefit onboard computers with limited resources so that deployed AI
no longer consumes precious computing resources to compute unnecessary actions,
effectively improving general system deployability.
In our approach, the policy determines the following action and the duration of
the next time step, making the entire learning process and applying policies
\emph{adaptive} to the specific demands of a given task. This paradigm shift
follows the core principles of reactive programming \citep{bregu2016reactive}:
as illustrated in the lower portion of Figure \ref{fig:overview}, in stark
contrast to a strategy reliant on fixed execution times, adopting a dynamic
execution time-based approach empowers the agent to achieve significant savings
in terms of computational resources, energy consumption, and time expended.
Moreover, our adaptive approach enables the integration of learning and control
strategies, resulting in a unified system that enhances data efficiency and
simplifies the pursuit of an optimal control strategy.

An immediate benefit of our approach is that the freed computational
resources can be allocated to additional tasks, such as perception and
communication, broadening the scope of RL applicability in
resource-constrained robots. We view the variable control rate
as promising for widely adopting RL in robotics.

\section{Reinforcement Learning with A Fixed Control Rate}

Before delving into the variable control rate RL, we provide a concise overview
of fixed time step-based RL. A notable example of successful real-world
reinforcement learning applications is Sony's autonomous racing car: Sony has
effectively harnessed the synergy between reinforcement learning algorithms
and a foundation of dynamic model knowledge to train AI racers that surpass
human capabilities, resulting in remarkably impressive performance outcomes
\citep{wurman2022outracing}. From a theoretical perspective,
\citet{li2020adaptive} aimed to bolster the robustness of RL within non-linear
systems, substantiating their advancements through simulations in a vehicular
context. A shared characteristic among these studies is their dependence on a
consistent control rate, typically 10 or 60~Hz. However, it is important to
note that a successful strategy does not necessarily equate to an optimal one.
As previously mentioned, time is critical in determining the system's
performance, whether viewed from an application or theory perspective. The
energy and time costs of completing a specific task determine an agent's level
of general efficiency. A superior control strategy should minimize the
presence of invalid instructions and ensure control actions are executed only
when necessary. Hence, the duration of an individual action step should not be
rigidly fixed; instead, it should vary based on the dynamic demands of the
task.

We have designed a variable control rate reinforcement learning method to address 
the time issue. This approach trains the policy to predict the time for executing 
the action and the action itself, adapting to dynamic changes in demand. The goal
is to overcome dynamic environment uncertainty and save the agent's computing 
resources. In a related study on changes in action execution time using 
reinforcement learning. \citet{sharma2017learning} introduced the concept of a 
learning-to-repeat strategy. Essentially, they found that when an agent performs 
the same action with the same execution time repeatedly, it can effectively alter 
the action execution time. However, their approach has limitations, particularly 
in simulating real-world scenarios. In simulations, executing a 10-second action 
and executing a 1-second action ten times may appear similar. Yet, their method 
doesn't account for potential changes in the agent's physical properties in 
natural environments. For instance, if the mass of the agent changes while it's 
performing a task, executing the same action simultaneously may not yield the 
same results. Moreover, the repeated execution of the same actions doesn't 
optimize computing resources. Allocating these resources to other tasks, like 
environment awareness, could significantly benefit onboard computers with limited 
computing capacity.

In addition, we also noticed that \citet{chen2021varlenmarl} changes the "control 
rate" in disguise by taking actions such as sleeping. However, it still needs a 
fixed control frequency to scan the agent's status to decide whether to take 
sleep action as the next step. High-frequency system status scanning and 
calculation also cannot effectively reduce the computing load.
In addition to the previously mentioned scenarios, there exists a diverse
range of time-sensitive reinforcement learning tasks spanning various domains.
These tasks cover multiple fields, including robotics, electricity markets,
and many others \citep{nasiriany2022augmenting, pardo2018time,zhang2019deep,
yang2018hierarchical}. However, using a fixed control rate is
a common thread among these works and systems like robots, which
often lack ample computing resources, can struggle to maintain a high and
fixed control rate.
The algorithm we designed can solve the problem of insufficient computing 
resources caused by maintaining too high a frequency, and it can be applied 
widely. It can train almost all RL strategies based on continuous control. For 
example, the problem of vehicles moving in a Newtonian dynamic environment; the 
classic robot arm problem: one assumes that the output of the strategy set is a 
force set. Suppose you want to quickly grab a fragile object, such as an egg, 
without considering other interferences, such as the material of the robot arm. 
In that case, the grip strength of the robot arm and direction are essential, 
and the duration of the force is even more critical. Beside the robotic related 
application, it can also be applied on Real-Time Strategy Applications, such as 
AI in gaming applications (Trackmania, Genshin Impact, and etc.). Especially those 
application scenarios deployed on smartphones with limited computing resources.

\section{Reinforcement Learning with Variable Control Rate} \label{section:III}
A straightforward approach to variable time step duration is to monitor the
completion of each execution action and dispatching the subsequent command.
Indeed, in low-frequency control scenarios, there are typically no extensive
demands for information delay or the overall time required to accomplish the
task \citep[games like Go or Chess, ][]{silver2016mastering,
	silver2018general}. However, in applications like robotics or autonomous
driving, the required control frequency can vary from very high
\citep{hwangbo2017control,hester2013texplore,hester2012rtmba} to low depending
on the state of the system. Following reactive programming
principles\citep{bregu2016reactive}, to control the system only when necessary we
propose that the policy also outputs the duration of the current time step.
Reducing the overall number of time steps conserves computational resources,
reduces the agent's energy consumption, and enhances data efficiency.

Unfortunately, in most RL algorithms, such as Q learning \citep{watkins1992q}
and the policy gradient algorithm \citep{sutton1999policy}, there is no
concept of the action execution time, which is considered only in few works
\citep{ramstedt2019real, bouteiller2021reinforcement}. When control frequency
is taken into consideration, it is mostly related to specific control problems
\citep{adam2011experience, almasi2020robust}, and assuming actions executed at
a fixed rate.

We propose a reward policy incorporating the agent's energy consumption and
the time taken to complete a task and extend Soft
Actor-Critic~\cite{haarnoja2018soft1} into the Soft Elastic
Actor-Critic (SEAC) algorithm, detailed in the following.

It is worth noting that for using SEAC on a robot control, an appropriate control 
system would be necessary for a real-world implementation, 
e.g. a proportional-integral-derivative controller (PID) controller
\citep{singh2013system}, an Extended Kalman filter (EKF)\citep{dai2019ekf}, and 
other essential elements. These components would need to be implemented to use 
SEAC into a real system environment. Nevertheless, we show that our system can 
indeed learn the duration of control steps and outperform established methods in 
a proof-of-concept implementation.

\subsection{Energy and Time Concerned Reward Policy}
\begin{figure}[h]
	\begin{center}
		\includegraphics[width=1.0\linewidth]{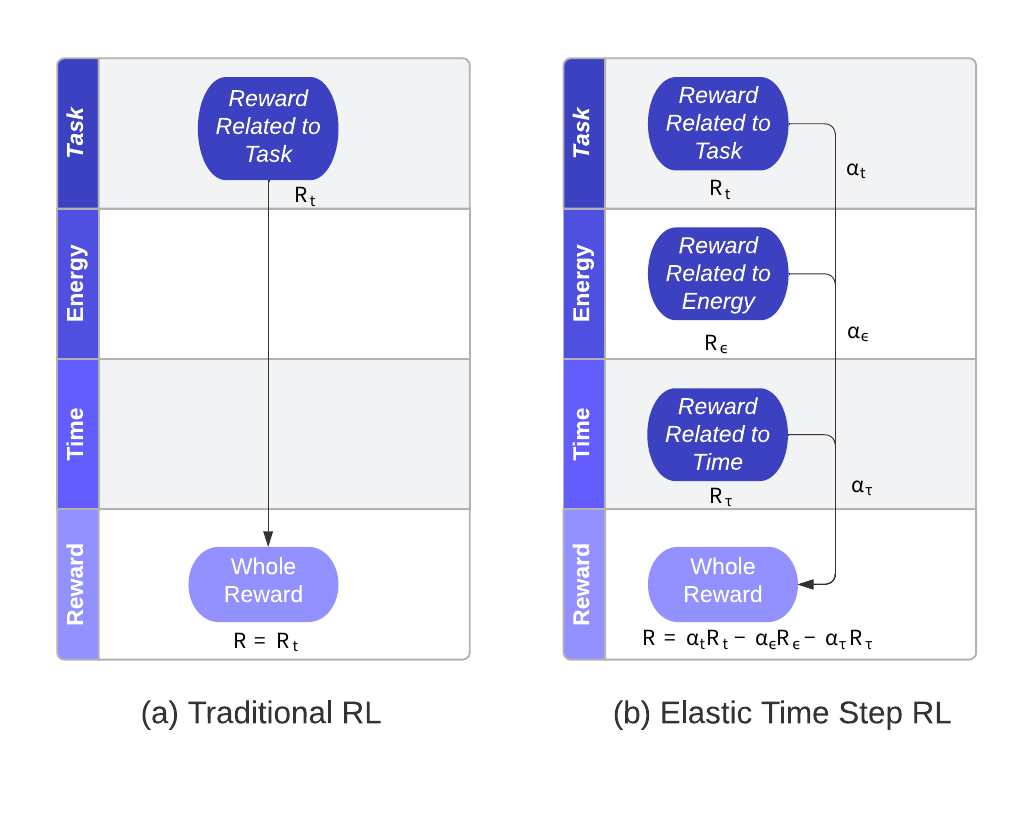}
	\end{center}
	\caption{(a) the reward policy for traditional RL; (b) the reward policy for elastic time step RL}
	\label{fig:reward_policy}
\end{figure}

As shown in Figure \ref{fig:reward_policy}, our approach tackles a multi-objective
optimization challenge, in contrast to conventional single-objective
reinforcement learning reward strategies. We aim to achieve a predefined
objective (metric 1) while minimizing energy consumption (metric 2) and time to
complete the task (metric 3). To reduce the reward to a scalar, we introduce 3
weighting factors: $\alpha_t$, $\alpha_{\varepsilon}$, and $\alpha_{\tau}$,
respectively assigned to our three metrics. It is important to note that we
consider only the energy consumption associated with the computation of a time
step (i.e. energy is linearly proportional to the number of steps) and not the
energy consumption of the action itself (e.g moving a heavy object, taking a
picture, etc.). Thus, our assessment of the agent's energy usage is solely based
on the \emph{computational load}.

At the same time, it is also necessary to minimize time consumption. Reducing 
the number of time steps does not necessarily mean reducing the overall time 
consumption under variable control rates. The MDP does not constrain the time 
the agent takes to complete a time step. In some extreme cases, for example, the 
agent took \emph{an entire year} to meet \emph{one step} only. Although it meets 
with the MDP, it is unacceptable for most application scenarios. Based on the 
above considerations, while we pursue the number of time-step minimization, we 
also pursue time-consuming minimization.

We assume that each action incurs a uniform energy consumption, denoted as
$\varepsilon$. The time taken to execute an action is $\tau$. In this context, 
the reward for one single step is represented by $R$, and the relationship can 
be expressed as follows:

\begin{myBox}[]{}{Reward Function}
	\begin{definition}\label{def:reward_function}
		The reward function is defined as:
		\begin{equation}
		R = \alpha_{t} \cdot R_t - \alpha_{\varepsilon} \cdot R_{\varepsilon} - 
		\alpha_{\tau} \cdot R_{\tau}
		\end{equation}
		where \\$R_t$ = $r$, r is the value determined by task setting;
		\\$R_{\varepsilon} =\varepsilon$,  $\varepsilon$ is the 
		energy cost of one time step;\\
		$R_{\tau} =\tau$, 	$\tau$ is the time taken to execute a time step.\\
	    $\alpha_t, \alpha_{\varepsilon}, \alpha_{\tau}$ are parametric weighting factors.
		
		We determine the optimal policy $\pi^*$, which maximizes the reward $R$.
	\end{definition}
\end{myBox}



While our algorithm aims to optimize multiple objectives, we take a simplified 
approach using a weighted strategy. Unlike the Hierarchical Reinforcement Learning 
(HRL) algorithm \citep{dietterich2000hierarchical, li2019hierarchical}, our 
method doesn't pursue Pareto optimality 
\citep{kacem2002pareto, monfared2021pareto} or involve multiple reward policies 
at different levels. Although our approach and HRL are designed to achieve various 
goals, they differ fundamentally in strategy settings. We've intentionally kept 
our strategy straightforward, making it user-friendly, computationally friendly, 
and adaptable for other algorithms. We've incorporated our reward policy into 
the SAC \citep{haarnoja2018soft1,haarnoja2018soft2} algorithm, extending it to 
SEAC. Importantly, this policy isn't exclusive to SAC – it can also be applied to 
other RL algorithms, such as PPO \citep{schulman2017proximal}, TD3 
\citep{fujimoto2018addressing}, etc. This compatibility makes our algorithm easily 
referenceable and deployable to different applications.

\subsection{The Verification Environment Design} \label{section:3_2}
Our SEAC verification work requires an environment that is non-discrete time and 
can reflect the full impact of time on action execution. This environment is not 
available within existing RL environments, we establish a test environment based 
on Gymnasium featuring variable action execution times, shown in
Figure \ref{fig:map}:

\begin{figure}[h]
	\begin{center}
		\includegraphics[width=0.8\linewidth]{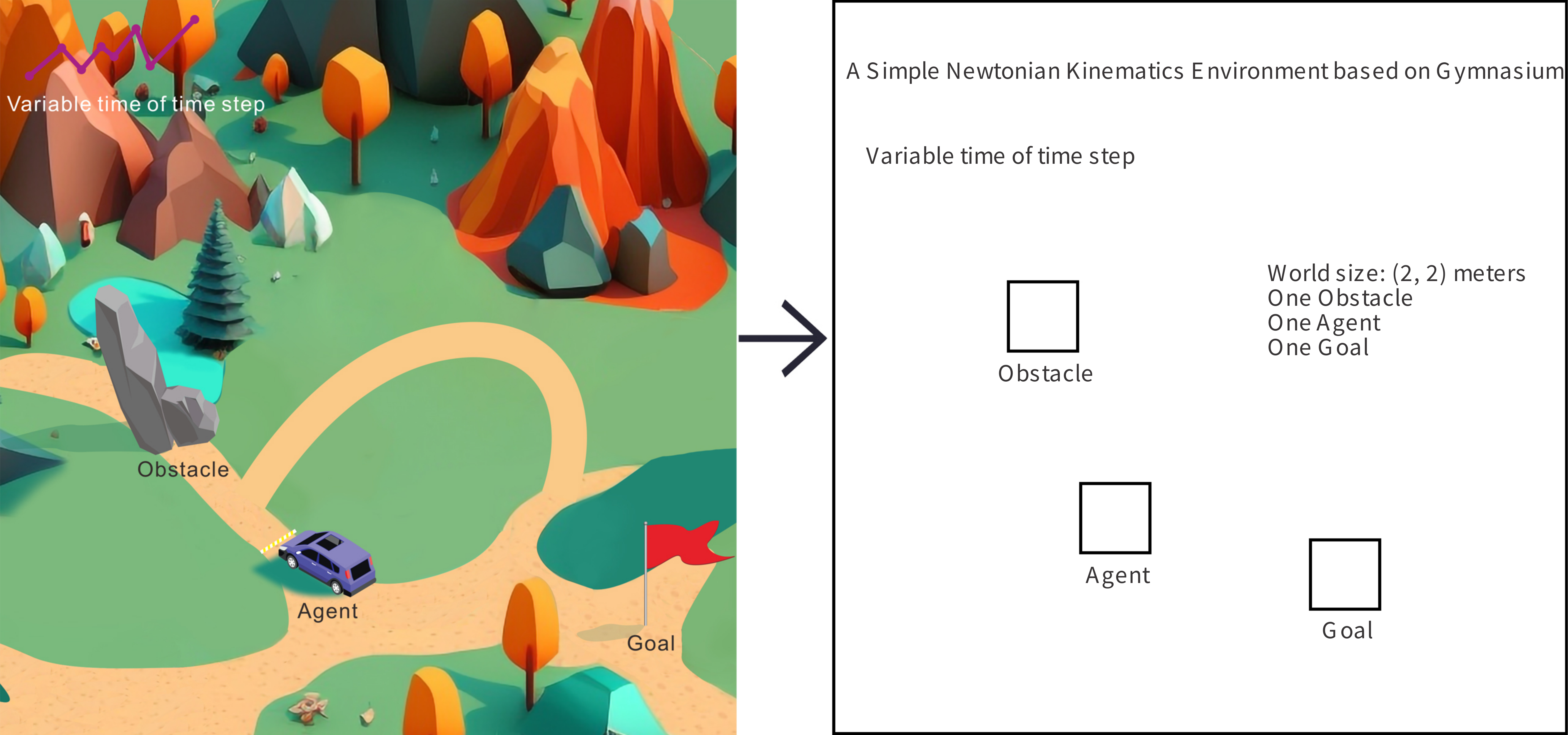}
	\end{center}
	\caption{A simple Newtonian Kinematics environment designed for verifying 
		SEAC based on gymnasium.}
	\label{fig:map}
\end{figure}

To help readers grasp our algorithm better, we will delve into the design of the 
verification environment and provide a detailed explanation of how we implemented 
the algorithm in the upcoming section. This will give a clearer picture of the 
practical aspects of our work.

This environment is a continuous two-dimensional (2D) and consists of a
starting point, a goal, and an obstacle. The task involves guiding an agent
from the starting point to the goal while avoiding the obstacle. Upon
resetting the environment, a new goal and obstacle are randomly generated. The
conclusion of an epode is reached when the agent reaches the goal or
encounters an obstacle. The agent is governed by Newton's laws of motion,
including friction, details can be found in Defination \ref{def:newton_formulas} 
and Appendix A \ref{Appendix:A}.

The starting point of the agent is also randomly determined. If the goal
or obstacle happen to be too close to the starting point, they are reset.
Similarly, if the goal is too close to the location where the obstacle was
generated, the obstacle's position is reset. This process continues until all
three points are situated at least 0.05 meters apart from each other on a 
(2 x 2) meters map. Meanwhile, the maximum force in a single step is 100.0 
Newton.

There are eleven dimensions of the state in the environment: the agent's
position, the position of the obstacle, the position of the goal, the speed of 
the agent, the duration for last time step, and the force being applied for the 
last time step. It is worth noting that we are indeed using historical data (i.e. 
the movement and duration value of the preceding step), but refrain from using 
recurrent neural networks (RNN) \citep{zaremba2014recurrent, lipton2015critical}. 
This decision stems from our concern that adopting recurrent architectures might 
deviate the overall reinforcement learning process from the Markov assumption 
\citep{norris1998markov, gers2000learning}: different decisions could arise from 
the same state due to the dynamic environment. 

We consider 3 dimensions to the actions within the environment:
\begin{compactenum}
	\item The time taken by the agent to execute the action;
	\item the force being applied for the agent along the x axis;
	\item the force being applied for the agent along the y axis.
\end{compactenum}

For instance, an action $a_t = (0.2, 50.0, -70.0)$ denotes that the agent is
expected to apply 50.0 Newton force along the x axis and -70.0 Newton force 
along the y axis within 0.2 seconds. For more detailed environment settings, 
see Appendix A \ref{Appendix:A}.

\section{Implementation of The SEAC}\label{section:IV}

We expend the SAC algorithm to verify the validity of our Definition 
\ref{def:reward_function}. Because we have not modified its loss functions, please 
refer to \citet{haarnoja2018soft2} for its policy optimization.

We validate our reward strategy by taking the SAC algorithm and implementing
fully connected neural networks \citep{muller1995neural} as both the actor and
critic strategy. We assume the agent can explore the unknown environment as
much as possible based on information entropy, giving a high probability that
the agent can discover the optimal solution to complete the task.

\begin{figure}[h]
	\begin{center}
		\includegraphics[width=1.0\linewidth]{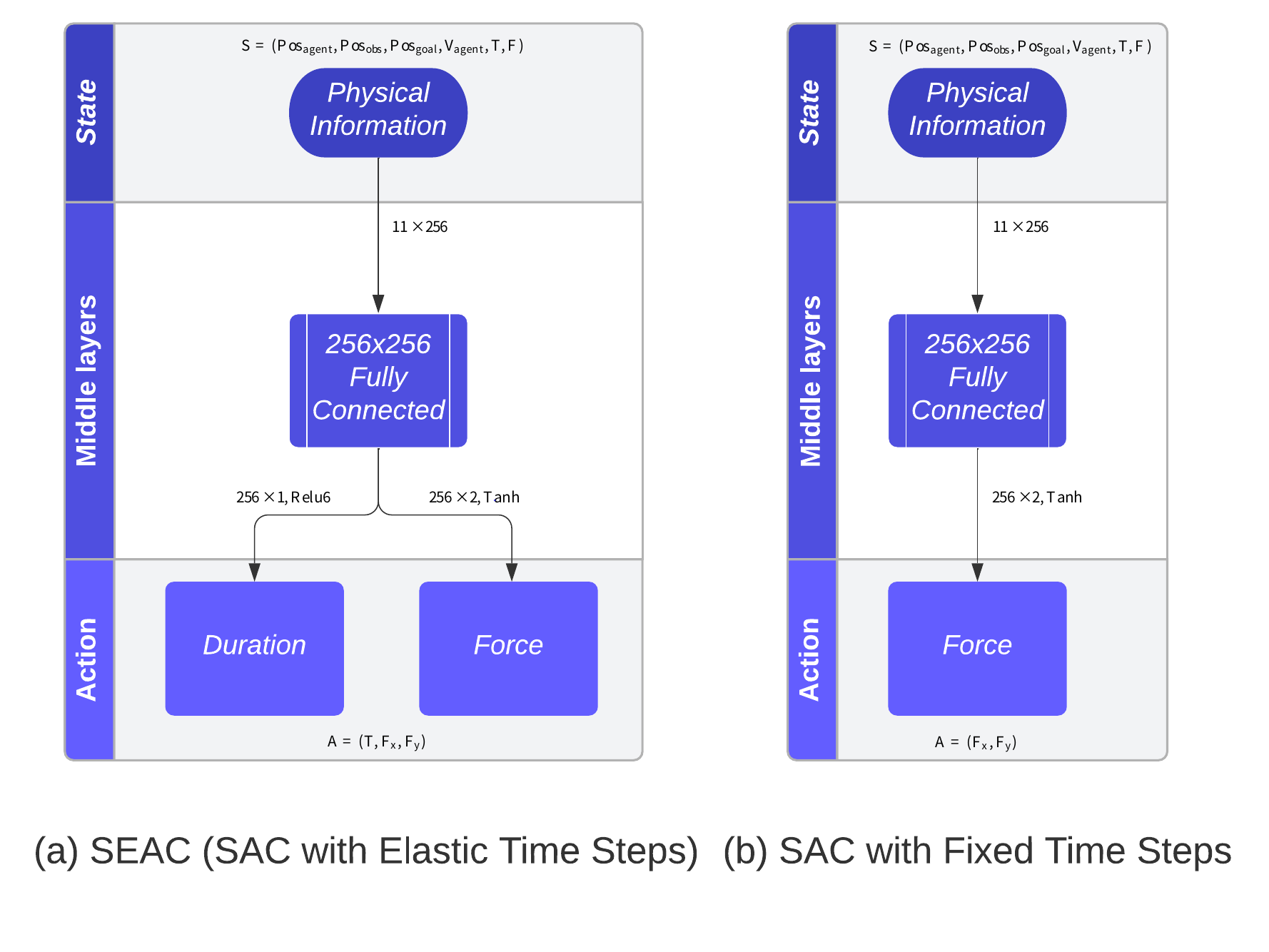}
	\end{center}
	\caption{(a) The SEAC ActorNetwork Architecture; (b) The SAC ActorNetwork Architecture.}
	\label{fig:architecture}
\end{figure}

As shown in Figure \ref{fig:architecture}, we marked the specific values of 
$S_{t}$ and $A_{t}$ based on the environment we explained in Section 3-2 
\ref{section:3_2}, and the network structure of the Actor Policy. This will 
intuitively help readers realize the difference between SEAC and SAC. In contrast 
to the conventional RL, in addition to the environment's state data, we take the 
agent's action history values and bring them together to the form of states as 
the network's input, including the time spent performing the previous action 
($T$), and the force in the last step ($F_{x}$ and $F_{y}$). We regrad them as 
part of the state, because we believe position and velocity is insufficient to 
express the system's inertia. This setting can ensure that the Markov process 
can completely describe our environment, thereby ensuring the convergence of 
the RL algorithm. Additionally, our approach involves an extra component at the 
output: the duration time $T$ for each action. 

When the Actor Network generates the action value $A_{t}$ for the next step, 
the controller (Figure \ref{fig:architecture}) will computes a range of 
control-related parameters (e.g., speed, acceleration, etc.) based on the action 
value and time. Under the context of our test environment, The related formulas 
are Newton Kinematics:

\begin{myBox}[]{}{Newton Kinematics Formulas}
	\begin{definition}\label{def:newton_formulas}
		The Newton Kinematics formulas are:
		\begin{equation}
		D_{aim} = 1/2 \cdot (V_{aim}+V_{current}) \cdot T;
		\end{equation}
		\begin{equation}
		V_{aim} = V_{current} + AT;
		\end{equation}
		\begin{equation}
		F_{aim} = mA;
		\end{equation}
		\begin{equation}
		F_{true} = F_{aim} – f_{friction};
		\end{equation}
		\begin{equation}
		\begin{split}
		f_{friction} &=\mu mg, If F_{aim} > f_{friction} \\
		& \& V_{agent} \neq 0;\\
		&= F_{aim}, If F_{aim} \leq f_{friction} \\
		& \& V_{agent} = 0.
		\end{split}
		\end{equation}
		where $D_{aim}$ is the distance generated by the policy that the agent 
		needs to move. $V_{agent}$ is the speed of the agent. $T$ is the time 
		to complete the movement generated by the policy, $m$ is the mass of 
		the agent, $\mu$ is the friction coefficient, and $g$ is the 
		acceleration of gravity.		
	\end{definition}
\end{myBox}

Through the duration and force generated by the policy, we can know the 
acceleration required to complete the policy and then calculate the speed. 
Then, we can compute the movement. However, due to friction, the actual 
acceleration will be inconsistent with the aimed acceleration, and the movement 
of the agent will be affected by Newtonian kinematics. Ultimately, the agent 
incorporates these actionable parameters into the environment, generating a new 
state and reward. This process is iterated until the completion of the task.

Our objective is for the agent to learn the optimal execution time for each
step independently. We need to ensure that time is not considered as a
negative value. Consequently, diverging from the single $Tanh$
\citep{kalman1992tanh} output layer typical in traditional RL Actor Networks,
we separate the Actor Network's output layer into two segments: we use $Tanh$
as the output activation for the action value, and $ReLu6$
\citep{howard2017mobilenets} for the output activation related to the time
value.


As Definition \ref{def:reward_function}, the precise
reward configuration for our environment are outlined in Table \ref{table:tab1}.
The hyperparameters settings can be found in Appendix B\ref{Appendix:B}.

\begin{table}[htbp]
	\centering
	\caption{Reward Policy for The Simple Newtonian Kinematics Environment}
	\begin{tabular}{llcl}
		\toprule
		\multicolumn{4}{c}{Reward Policy} \\
		\midrule
		Name                  & Value                  &                 & Annotation                 \\
		\midrule
		& $100.0$                   &                 & Reach the goal             \\
		r                     & $-100.0$                  &                 & Crash on an obstacle       \\
		& $-1.0 \cdot D_{goal}$   &                &$D_{goal}$: distance to goal              \\
		$\epsilon$            & $1.0$      &     & Computational energy (Joule) \\
		$\alpha_t$            & $1.0$      &    & Task gain factor                    \\
		$\alpha_{\epsilon}$   & $1.0$      &    & Energy gain factor                    \\
		$\alpha_{\tau}$       & $1.0$      &    & Time gain factor                    \\
		\bottomrule
	\end{tabular}
	\label{table:tab1}
\end{table}

It is noteworthy that our design does not substantially increase the model size. 
The recorded size of the network model in Appendix A is a mere 282.5KB after 
training. In comparison, the model size of the SAC network, of equivalent 
dimensions, is 280.5KB after training. The SEAC model is only 2KB larger, 
representing approximately a \emph{7 \textperthousand} increase in storage demand. 
This meticulous consideration ensures the model's deployability.

\section{Experimental Results} \label{section:V}

\begin{figure}[h]
	\begin{center}
		\includegraphics[width=1.0\linewidth]{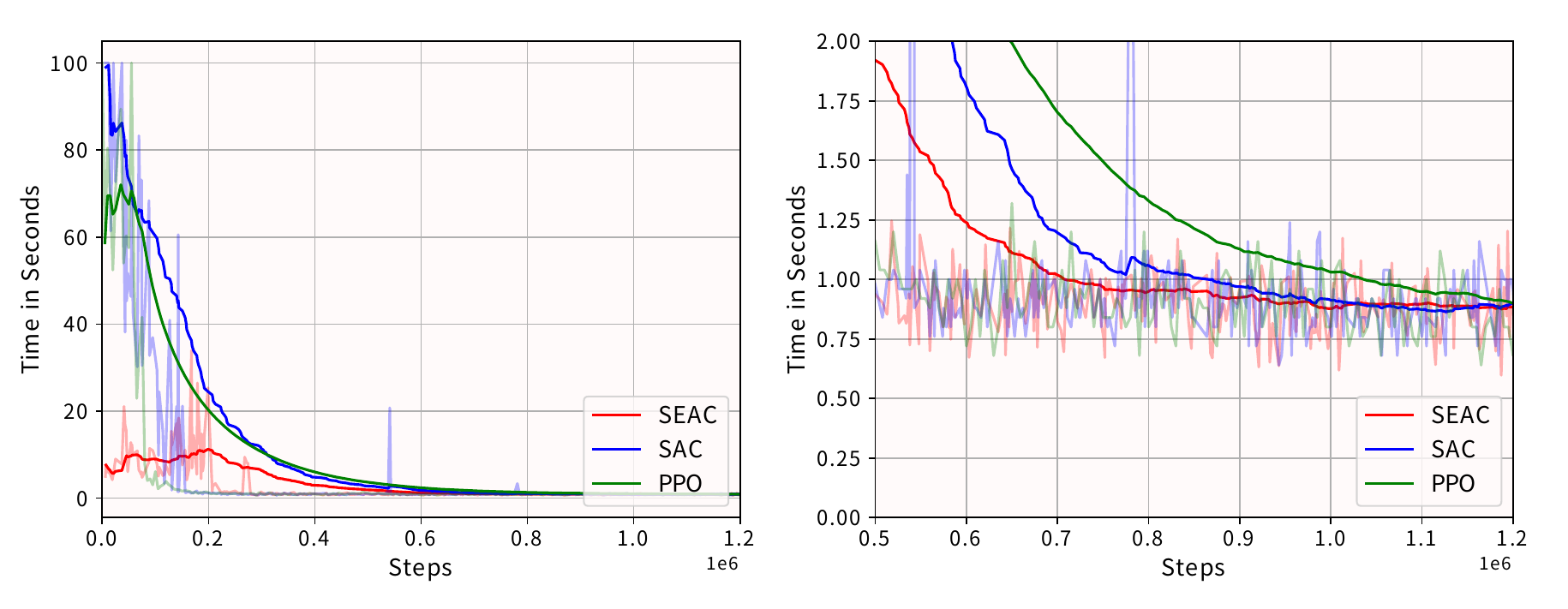}
	\end{center}
	\caption{Average time cost per episode for three algorithms trained in 1.2 
		millions steps. The figure on the right is a partially enlarged version 
		of the figure on the left.}
	\label{fig:average_time_cost}
\end{figure}

\begin{figure}[h]
	\begin{center}
		\includegraphics[width=1.0\linewidth]{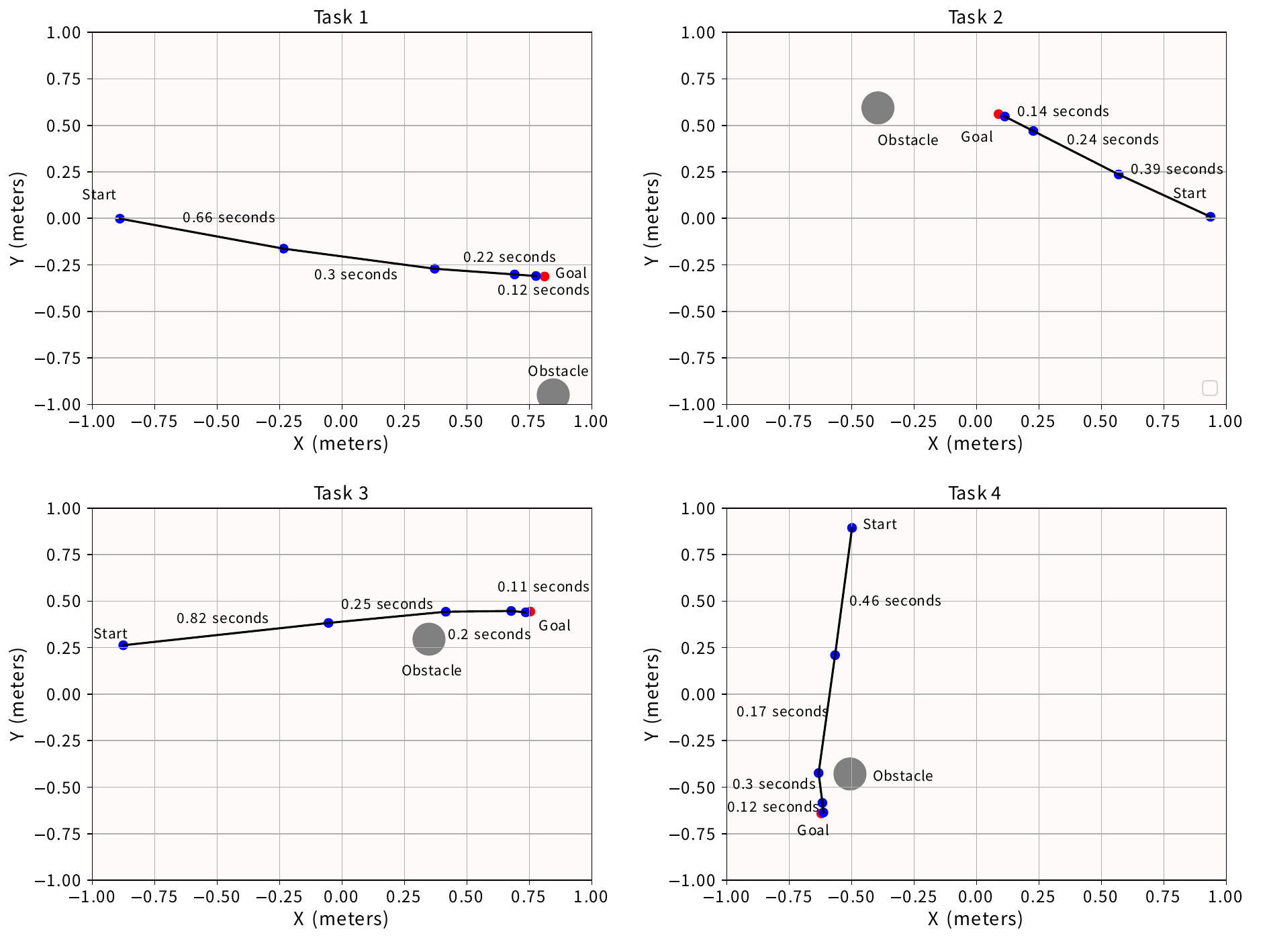}
	\end{center}
	\caption{Four example tasks show how SEAC changes the control rate 
		dynamically to adapt to the Newtonian mechanics environment and 
		ultimately reasonably complete the goal.}
	\label{fig:frequency_graph}
\end{figure}

\begin{figure}[h]
	\begin{center}
		\includegraphics[width=1.0\linewidth]{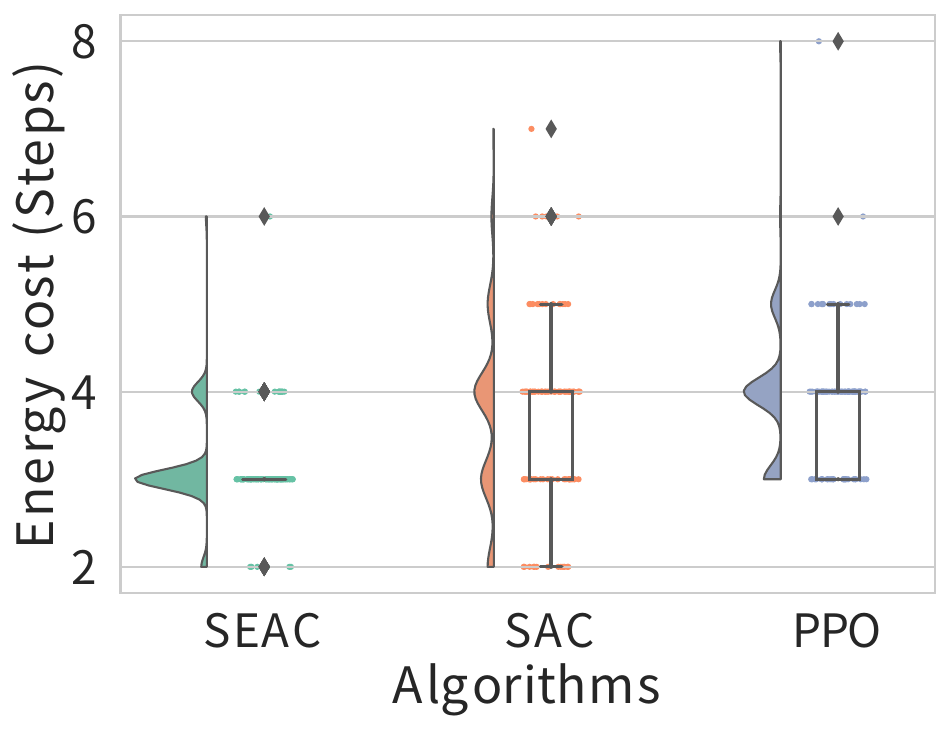}
	\end{center}
	\caption{The energy cost for 100 trials. SEAC consistently reduces the 
		number of time steps compared with PPO and SAC without affecting the 
		overall average reward. Therefore, SAC and PPO are not optimizing for 
		energy consumption and have a much larger spread.}
	\label{fig:energy_graph}
\end{figure}

We conducted six experiments for each of the three RL algorithms, employing
various parameters within the environment described in Section III\ref{section:3_2}
\footnote{Our code is publicly available at Github: 
	https://github.com/alpaficia/SEAC\_Pytorch\_release}. These experiments 
were conducted on a machine equipped with an
Intel Core i5-13600K CPU and an NVIDIA RTX 4070 GPU, running Ubuntu 20.04.
Subsequently, we selected the best-performing policy for each of these three
algorithms to draw the graphs in Figures \ref{fig:average_returns}--\ref{fig:energy_graph}.

The frequency range for action execution spans from 1 to 100 Hz, and the
agent's force value ranges from -100 to 100 Newton. We compared our
results with the original SAC \citep{haarnoja2018soft2} and PPO
\citep{schulman2017proximal} algorithms, both employing a fixed action
execution frequency of 5.0~Hz. 

We use the conventional average return graph and record the average time cost 
per task to clearly and intuitively represent our approach's performance. 
Furthermore, we generate a chart illustrating how the variable control rate policy 
works in action execution frequency for four tasks with the SEAC model. Finally, 
we draw a raincloud graph to visualize the disparities in energy costs among 
these three RL algorithms for one hundred different tasks.

Appendix B\ref{Appendix:B} provides all hyperparameter settings and implementation
details. The average return results of all algorithms are shown in
Figure \ref{fig:average_returns}, and their time-consuming results are shown in
Figure \ref{fig:average_time_cost}:

\begin{figure}[h]
	\begin{center}
		\includegraphics[width=1.0\linewidth]{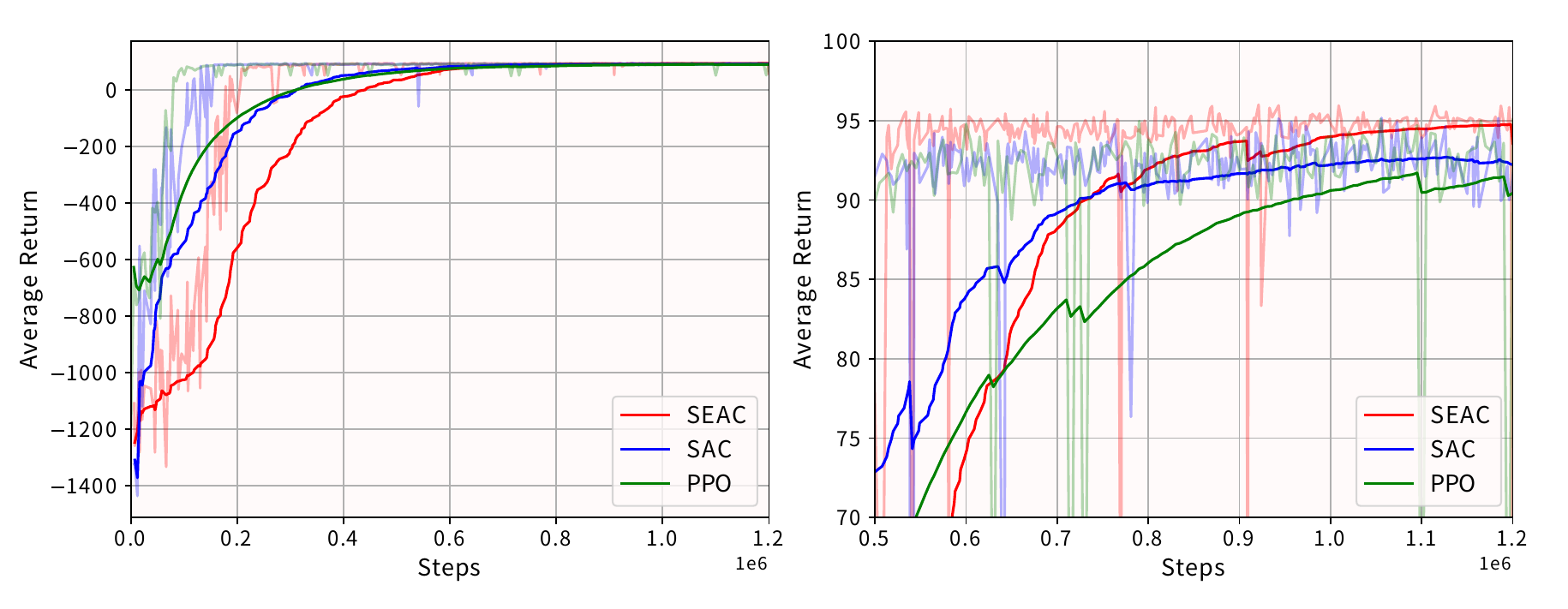}
	\end{center}
	\caption{Average returns for three algorithms trained in 1.2 million steps. 
		The figure on the right is a partially enlarged version of the figure on 
		the left.}
	\label{fig:average_returns}
\end{figure}

Figure \ref{fig:average_returns} and Figure \ref{fig:average_time_cost} show that
SEAC surpasses the baselines in terms of average return and time efficiency. 
The amount of data that SEAC needs to be trained has one more dimension than SAC
 and PPO (duration). So its training speed is not as fast as the above two at 
 the beginning. However, when SEAC gradually reduces the amount of data by 
 reducing the number of steps, its data efficiency improves 
 significantly, starting from around 300,000 steps and finally converging before 
 SAC and PPO around 900,000 to 1,200,000 steps. Fast training is also an 
 exciting advantage for deployable AI.Meanwhile, SEAC, using the same policy 
 optimization algorithm but incorporating an elastic time step, demonstrates 
 higher and more stable final performance than SAC.

When considering the adaptation of action execution frequency within the SEAC 
model, we generate the variable control rate policy explanation charts for four 
distinct trials, as depicted in Figure \ref{fig:frequency_graph}, each utilizing 
different random seeds. As shown within the four charts, the first step of these 
SEAC policies is to use one or two \emph{long period} steps to give the agent a 
large force to accelerate and to approach the goal, then use \emph{a small time} 
to adjust the speed direction within a step, and then use high-frequency control 
rates to reach the target (subfigure 4) stably. Subfigures 3 and 4 show that 
it has learned to avoid obstacles. 

Additionally, Figure \ref{fig:energy_graph} illustrates the energy cost (i.e. the
number of time steps) across one hundred independent trial: as expected, SEAC
minimizes energy with respect to PPO and SAC without affecting the overall
average reward. It is worth noting that SAC and PPO are not optimising for
energy consumption, so they are expected to have a large result spread. More
interestingly, SEAC \emph{both} reduces energy consumption \emph{and} achieved
a high reward. We maintain a uniform seed for all algorithms during this
analysis to ensure fair and consistent results.

As shown in Figure \ref{fig:frequency_graph}, the agent's task execution strategy
primarily focuses on minimizing the number of steps and the time required to
complete the task. Notably, the agent often invests a substantial but
justifiable amount of time in the initial movement phase, followed by smaller
times for subsequent steps to arrive at the goal. This pattern aligns with our
core philosophy of minimizing energy and time consumption.

Figure \ref{fig:energy_graph} shows that the average computing energy 
consumption and its standard deviation of SEAC for completing the same task are 
far lower than the two fixed control rate-based reinforcement learning policies 
as comparison objects. Compared with the SAC baseline, the average computing 
energy consumption is reduced by arround 25\% while taking less time to 
accomplish tasks. As we mentioned in Section IV\ref{section:IV}, when the 
storage load increased by about 7 \textperthousand, the computing energy 
consumption dropped by around 25\%, and the performance was not lost or even 
better. Compared with the fixed-time step RL policy, the variable-rate RL policy 
is undoubtedly \emph{a better choice} for the deployable AI.

Furthermore, the variance in data distribution is notably reduced within the
SEAC results. These findings underscore the algorithm's heightened stability
in dynamic environments, further substantiating the practicality of our
variable control rate-based RL reward policy.

\section{Conclusions and Future Work}

We propose a variable control rate-based reward policy that allows an agent to
decide the duration of a time step in reinforcement learning, reducing energy
consumption and increasing sample efficiency (since fewer time steps are
needed to reach a goal). Reducing the number of time steps can be very
beneficial when using robots with limited capabilities, as the newly freed
computational resources can be used for other tasks such as perception,
communication, or mapping. The overall energy reduction also increases the 
applicability of this policy to deployable AI on robotics.

We introduce the Soft Elastic Actor Critic (SEAC) algorithm and verify its
applicability with a proof-of-concept implementation in an environment with
Newtonian kinematics. The algorithm could be easily extended to real-world
applications, and we invite the reader to refer to Section V\ref{section:V} and
Appendix B\ref{Appendix:B} for the implementation details.

To the best of our knowledge, SEAC is the first reinforcement learning
algorithm that simultaneously outputs actions and the duration of the
following time step. Although the method would benefit from testing in more
realistic and dynamic settings, such as Mujoco \citep{todorov2012mujoco} or TMRL
\citep{tmrl}, 
we believe this method presents a promising approach to enhance RL's efficiency 
and energy conservation for deployment on real-world robotic systems.

\bibliography{aaai24}

\appendix
\clearpage
\section{Appendix:A Validation Environment Details} \label{Appendix:A}

The Spatial Information of our environment are:
\begin{table}[htbp]
	\centering
	\caption{Details of The Simple Newtonian Kinematics Gymnasium Environment}
	\begin{tabular}{llcl}
		\toprule
		\multicolumn{4}{c}{Environment details} \\
		\midrule
		Name                  & Value                  &                 & Annotation                 \\
		\midrule
		Action dimension     & $3$                   &                 & \\
		Range of speed       & $[-2, 2]$                  &                 &m/s   \\
		Action Space         & $[-100.0, 100.0]$   &                &Newton  \\
		Range of time        & $[0.01, 1.0]$      &     &second  \\
		State dimension      & $6$      &    & Task gain factor                    \\
		World size           & $(2.0, 2.0)$      &    & in meters                    \\
		Obstacle shape       & Round      &    & Radius: 5cm                   \\
		Agent weight         & 20      &    & in $Kg$                   \\
		Gravity factor       & 9.80665      &    & in $m/s^2$                   \\
		Static friction coeddicient      & 0.6      &    &                   \\
		\bottomrule
	\end{tabular}
	\label{table:tab2}
\end{table}

The state dimensions are:

\begin{myBox}[]{}{State}
	\begin{definition}\label{def:state}
		State
		
		We have only one starting point, endpoint, and obstacle in the set environment. The positions of the endpoint and obstacle are randomized in each episode.\\
		\begin{equation}
		S_t=(Pos, Obs, Goal, Speed, Time, Force)
		\end{equation}
		
		Where:\\
		$Pos$ = Position Data of The Agent in X and Y Direction, \\
		$Obs$ = Position Data of The Obstacle in X and Y Direction, \\
		$Goal$ = Position Data of The Goal in X and Y Direction,\\
		$Speed$ = Spped Data of The Agent for Current Step,\\
		$Time$ = The Duration data of The Agent to Execute The Last Time Step,\\
		$Force$ = Force Data of The Agent for Last Time Step in X and Y Direction.
	\end{definition}
\end{myBox}

The action dimensions are:

\begin{myBox}[]{}{Action}\label{def:action}
	\begin{definition}
		Action
		
		Each action should have a corresponding execution duration.\\
		\begin{equation}
		a_t=(a_{t}, a_{fx}, a_{fy})
		\end{equation}
		
		Where:\\
		$a_{t}$ = The time of The Agent to implement current action, \\
		$a_{fx}$ = The Force of The Agent in X Coordinate, \\
		$a_{fy}$ = The Force of The Agent in Y Coordinate.
	\end{definition}
\end{myBox}

\clearpage
\section{Appendix:B Hyperparameter Setting of SEAC} \label{Appendix:B}	
\begin{table}[htbp]
	\centering
	\caption{Hyperparameters Setting of SEAC}
	\begin{tabular}{llcl}
		\toprule
		\multicolumn{4}{c}{Hyperparameter sheet} \\
		\midrule
		Name                  & Value                  &                 & Annotation                 \\
		\midrule
		Total steps       & $3e6$                   &                 &  \\
		$\gamma$          & $0.99$                  &          & Discount factor  \\
		Net shape         & $(256, 256)$   &                &  \\
		batch\_size       & $256$      &     &  \\
		a\_lr             & $2e4$      &    & Learning rate of Actor Network                    \\
		c\_lr             & $2e4$      &    & Learning rate of Critic Network      \\
		max\_steps        & $500$      &    & Maximum steps for one episode                   \\
		$\alpha$          & $0.12$      &    &                    \\
		$\eta$            & $-3$      &    & Refer to SAC \citep{haarnoja2018soft2}                   \\
		$T_i$             & $5.0$      &    & Time duration of rest compared RL algorithms, in HZ   \\
		min\_frequency    & $1.0$      &    & Minimum control frequency, in HZ                   \\
		max\_frequency    & $100.0$      &    & Maximum control frequency, in HZ                \\
		Optimizer         & Adam      &    & Refer to Adam \citep{kingma2014adam}                   \\
		environment steps & $1$       &    &              \\
		Replaybuffer size & $1e6$       &    &              \\
		Number of samples before training start & $5 \cdot max\_steps$       &    &              \\
		Number of critics & $2$       &    &              \\
		\bottomrule
	\end{tabular}
	\label{table:tab3}
\end{table}

\end{document}